\documentclass{article}
\usepackage{spconf,amsmath,graphicx,hyperref}
\usepackage{amssymb}
\usepackage{booktabs}
\usepackage{textcomp}

\newcommand{\best}[1]{\leavevmode\ooalign{#1\cr\hidewidth\kern0.08pt #1\hidewidth\cr}}

\title{MagicGarment: Benchmarking Single-Image Standalone 3D Garment Reconstruction}

\name{Jia Chen$^{1,2}$ \quad Ruyang Zhao$^{1}$ \quad
Jun Bai$^{3^{*}}$\thanks{$^{*}$ Corresponding author: \texttt{junbai.research@gmail.com}.} \quad
Chao Huang$^{1}$ \quad Saishang Zhong$^{1}$ \quad Xinrong Hu$^{1}$}

\address{
{\fontsize{9}{10.8}\selectfont $^{1}$ School of Computer Science and Artificial Intelligence, Wuhan Textile University, Wuhan 430200, China}\\
{\fontsize{9}{10.8}\selectfont $^{2}$ Engineering Research Center of Hubei Province for Clothing Information, Wuhan Textile University, Wuhan 430200, China}\\
{\fontsize{9}{10.8}\selectfont $^{3}$ School of Computer Science, McGill University, Montreal, QC H3A 0E9, Canada}
}

\begin{document}
\ninept
\maketitle
\begin{abstract}

\noindent Single-image 3D reconstruction has advanced rapidly, yet standalone garments remain challenging due to thin-shell geometry, open boundaries, and layered structures. Existing datasets and evaluation protocols provide limited support for systematic evaluation in this setting. We introduce \textbf{MagicGarment}, a benchmark for single-image standalone 3D garment reconstruction with 1,518 professionally constructed assets across 17 composite garment types, aligned multimodal annotations, and a unified evaluation protocol. We further establish a garment-aware reference baseline by adapting a pretrained image-to-3D backbone at two levels: Grounded Semantic Refinement (GSR) provides garment-focused visual conditioning, while Flow-Time-Route LoRA (FTR-LoRA) adapts the frozen 3D generative backbone using flow-timestep, token-routing, and expert-load information. Experiments demonstrate MagicGarment's benchmark utility and the effectiveness of garment-aware adaptation. Project page: \url{https://1019538157.github.io/MagicGarment/}.

\end{abstract}
%

\section{Introduction}\label{sec:introduction}


Single-image 3D reconstruction has advanced rapidly through multi-view diffusion~\cite{Long2024Wonder3D}, large reconstruction models~\cite{Tang2024LGM}, and 3D generative priors~\cite{Xiang2025TRELLIS}. However, reconstructing standalone garments from a single image remains particularly challenging. Unlike general rigid objects or clothed humans, garments exhibit thin-shell surfaces, open boundaries, hollow structures, wrinkles, and layered or multi-piece configurations. These properties conflict with generic watertight-object priors, often leading to incorrectly closed necklines, cuffs, and hems. Moreover, clothed-human reconstruction methods~\cite{Jiang2020BCNet,DeLuigi2023DrapeNet} typically model garments jointly with the underlying body, making them less suitable for standalone garment reconstruction.



Progress is further constrained by the lack of a unified benchmark for single-image standalone garment reconstruction. Existing datasets, including MGN~\cite{Bhatnagar2019MGN}, DeepFashion3D~\cite{Zhu2020DeepFashion3D}, CLOTH3D~\cite{Bertiche2020CLOTH3D}, 4D-DRESS~\cite{Wang2024FourDDress}, and GarVerseLOD~\cite{Luo2024GarVerseLOD}, support related tasks such as clothed-human reconstruction and garment modeling, but remain limited for this setting in garment-type coverage, stylized and multi-piece designs, and the consistency of geometry, texture, and multimodal annotations. Differences in task definitions and evaluation protocols further hinder systematic comparison between general-purpose and garment-specific reconstruction methods.

\begin{figure}[!th]
    \centering
    \includegraphics[width=\linewidth]{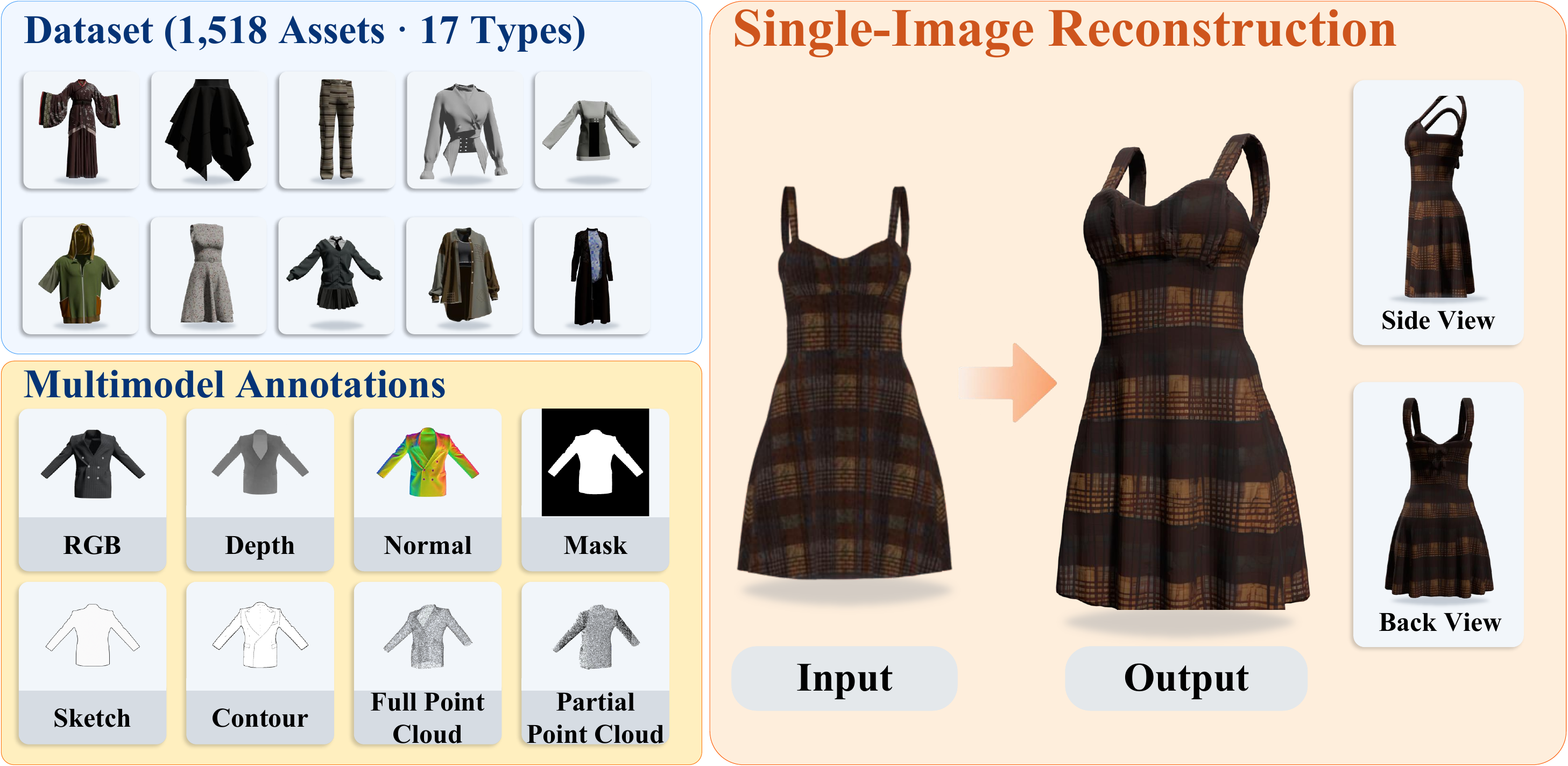}
    \caption{Overview of MagicGarment. Left: representative garment assets and aligned multimodal annotations. Right: the single-image standalone 3D garment reconstruction task.}
    \label{fig:magicgarment-overview}
\end{figure}

\begin{figure*}[t]
    \centering
    \includegraphics[width=.95\linewidth]{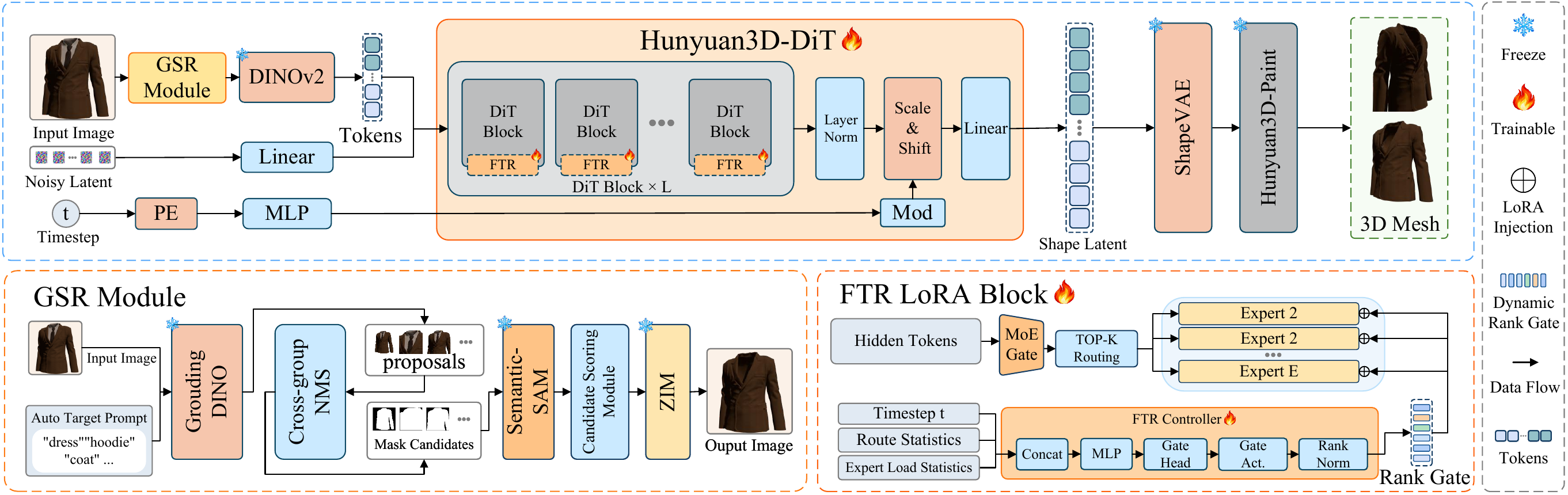}
    \caption{Overview of the garment-aware reference baseline built on pretrained Hunyuan3D. GSR performs input-level garment refinement, while FTR-LoRA provides model-level adaptation of the frozen Hunyuan3D-DiT backbone.}
    \label{fig:method-overview}
\end{figure*}

\begin{figure*}[t]
    \centering
    \includegraphics[width=.90\linewidth]{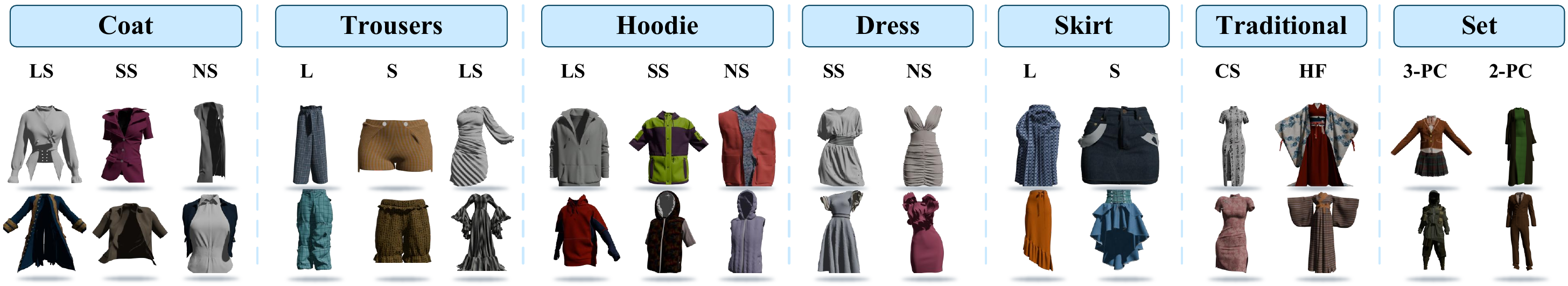}
    \caption{Representative 3D garment assets across the 17 composite types in MagicGarment.}
    \label{fig:magicgarment-examples}
\end{figure*}

To address this gap, we introduce \textbf{MagicGarment}, a benchmark for single-image standalone 3D garment reconstruction as illustrated in Fig.~\ref{fig:magicgarment-overview}. It contains 1,518 professionally constructed garment assets across 17 composite garment types, with aligned RGB, mask, depth, normal, text, point-cloud, and 2D-pattern annotations. We further establish a unified benchmark protocol covering geometric accuracy, appearance fidelity, surface orientation, and cross-modal semantic consistency. To provide a strong reference baseline, we adapt a pretrained image-to-3D backbone at two complementary levels: Grounded Semantic Refinement (GSR) refines garment-focused input conditioning by suppressing human and background interference, while Flow-Time-Route LoRA (FTR-LoRA) transfers the frozen image-to-3D backbone with strong 3D generative priors to the garment domain.





The main contributions of this work are threefold:
\begin{itemize}
    \item We introduce \textbf{MagicGarment}, a benchmark for single-image standalone 3D garment reconstruction, comprising 1,518 professionally constructed garment assets across 17 composite garment types, aligned multimodal annotations, and a unified evaluation protocol.

    \item We develop a garment-aware reference baseline that transfers a general-purpose image-to-3D backbone with strong 3D generative priors to the garment domain through GSR and FTR-LoRA at the input and model levels, respectively.

    \item We systematically benchmark representative garment-\allowbreak{}specific and general-purpose 3D reconstruction methods under a common protocol, providing evaluation across geometry, appearance, surface orientation, and semantic consistency.
\end{itemize}



\section{MagicGarment Benchmark and Reference Baseline}
\label{sec:benchmark_method}


\begin{table}[t]
\caption{Comparison of representative 3D garment datasets.}
\label{tab:dataset-quality-comparison}
\centering
\setlength{\tabcolsep}{0.2pt}
\begin{tabular*}{\linewidth}{@{}l@{\extracolsep{\fill}}ccccc@{}}
\toprule
Dataset
& \shortstack{Avg.\\\#Vertices}
& \shortstack{Avg.\\\#Faces}
& \#Types
& Method
& \#Annotations \\
\midrule
MagicGarment & \textbf{375,817} & \textbf{665,488} & \textbf{17} & \begin{tabular}[t]{@{}c@{}}Professional\\3D artist\\modeling\end{tabular} & 9 \\
SewFactory & 8,813 & 16,834 & 13 & Simulation & 3 \\
CLOTH3D & 7,473 & 14,329 & 6 & Simulation & 3 \\
DeepFashion3D & 21,782 & 42,613 & 10 & Real scans & 4 \\
GarmentCode v2 & 14,035 & 26,384 & N/A & Simulation & 4 \\
\bottomrule
\end{tabular*}
\end{table}




\subsection{MagicGarment Benchmark}
\label{subsec:benchmark}

\textbf{Dataset Construction.}
MagicGarment contains 1,518 standalone 3D garment assets across 17 composite garment types, with representative examples shown in Fig.~\ref{fig:magicgarment-examples}. All assets are professionally constructed using CLO~\cite{CLO3D2026}. Specifically, 500 human body models with diverse body shapes and poses are used as supporting forms for garment design and cloth simulation. Professional 3D garment artists create 2D patterns, specify sewing relationships and fabric properties, and perform physically based draping simulation. Only the garment geometry is retained in the final assets, ensuring that the target representation is independent of the human body.

All exported assets undergo manual quality inspection. Samples with missing structures, severe self-intersections, abnormal interpenetrations, or unstable geometry are removed. The retained meshes are converted to a unified right-handed coordinate system and spatially normalized. As summarized in Table~\ref{tab:dataset-quality-comparison}, MagicGarment provides substantially denser garment geometry and broader garment-type coverage than several representative existing datasets.

\textbf{Multimodal Annotations.}
For each garment asset, MagicGarment provides aligned multi-view RGB images, masks, depth maps, normal maps, contours, sketches, textual descriptions, complete and partial point clouds, and 2D pattern-related information. These annotations jointly characterize garment appearance, geometry, structure, and semantics, enabling both reconstruction benchmarking and downstream garment understanding tasks.

\textbf{Benchmark Protocol.}
We define single-image standalone 3D garment reconstruction as the benchmark task. Given a single garment image, a method is required to reconstruct the corresponding standalone 3D garment without access to additional views at inference time. Training, validation, and test sets are split at the garment-asset level to prevent instance overlap. All reconstructed outputs are converted into evaluable meshes and processed using the same coordinate, scale, surface-sampling, and alignment procedures. Evaluation covers four aspects: geometric accuracy, appearance fidelity, surface orientation, and cross-modal semantic consistency.

\begin{figure*}[t]
    \centering
    \includegraphics[width=.90\linewidth]{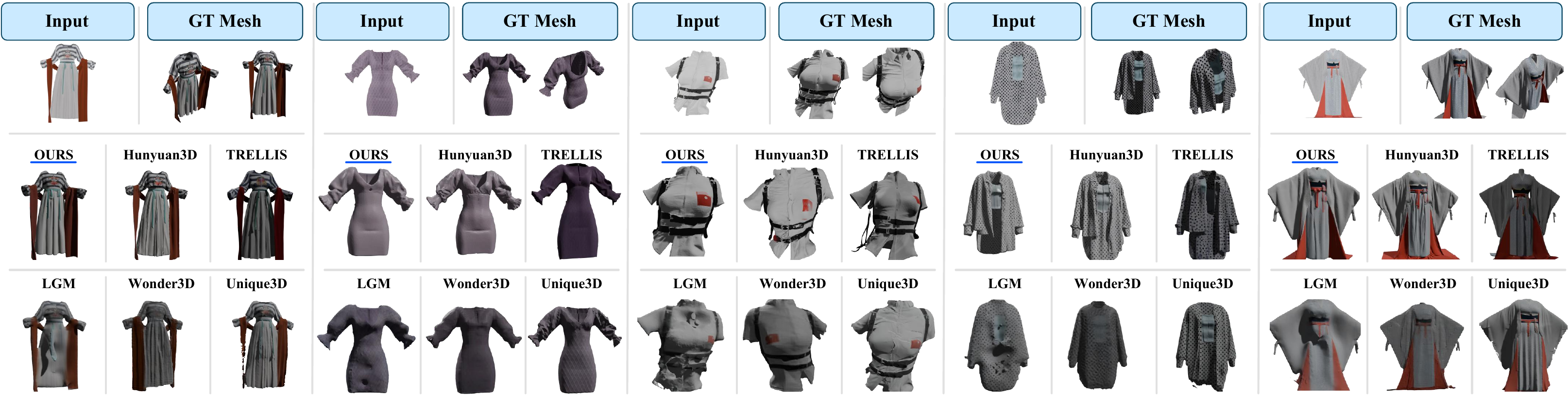}
    \caption{Qualitative comparison on MagicGarment. Each example shows the input image, ground-truth mesh, and reconstructions from our garment-aware baseline and representative image-to-3D methods.}
    \label{fig:qualitative-results}
\end{figure*}

\subsection{Garment-Aware Reference Baseline}
\label{subsec:baseline}

As illustrated in Fig.~\ref{fig:method-overview}, our reference baseline is built upon the pretrained Hunyuan3D image-to-3D framework. Given a single garment image, GSR first refines the visual input to obtain garment-focused conditioning, which is encoded into visual tokens by DINOv2. Together with the noisy shape latent and flow timestep, these tokens condition the Hunyuan3D-DiT backbone to generate a 3D shape latent. The resulting latent is decoded by ShapeVAE and further processed by Hunyuan3D-Paint to produce the final textured garment mesh. Within this pipeline, we introduce garment-specific adaptation at two complementary levels: GSR refines the input conditioning, while FTR-LoRA dynamically adapts the frozen Hunyuan3D-DiT backbone to garment-specific geometry.

\textbf{Grounded Semantic Refinement (GSR).}
Natural garment images often contain interference from the human body, background, and shadows, which can bias visual conditioning toward non-garment regions. GSR therefore extracts a garment-focused conditioning image before visual encoding. Given an input image and an automatically generated garment-category prompt, Grounding DINO~\cite{liu2024grounding} first localizes candidate garment regions, after which Semantic-SAM~\cite{Li2024SemanticSAM} generates multi-granularity mask candidates. Rather than relying on a single confidence score, GSR selects the foreground mask by jointly considering semantic consistency, segmentation stability, localization confidence, spatial consistency, and garment structural completeness:
\begin{equation}
i^\star=\arg\max_i
\left[
w_q q_i+w_u u_i+
w_b\psi_{\mathrm{box}}(C_i)+
w_\rho\rho_i+
w_s\psi_{\mathrm{str}}(C_i)
\right],
\label{eq:gsr_selection}
\end{equation}
where $C_i$ is the $i$-th mask candidate; $q_i$, $u_i$, and $\rho_i$ denote semantic consistency, segmentation stability, and localization confidence, respectively; and $\psi_{\mathrm{box}}$ and $\psi_{\mathrm{str}}$ measure localization consistency and garment structural plausibility.

The selected mask $M^\star=C_{i^\star}$ is further refined using ZIM (Zero-Shot Image Matting)~\cite{kim2025zim}, which produces fine-grained alpha mattes for detailed boundaries such as necklines, cuffs, hems, and cutouts. Among the refinement candidates $A_h$, the final alpha mask is selected as
\begin{equation}
h^\star=\arg\max_h
\left[
0.7\,\mathrm{IoU}\!\left(\mathbb{I}(A_h>\tau),M^\star\right)
+0.3\,\eta_h
\right],
\label{eq:gsr_refinement}
\end{equation}
where $\eta_h$ denotes the confidence of candidate $A_h$. The refined alpha mask is combined with the original RGB image to form garment-focused conditioning, reducing interference from non-garment content during 3D reconstruction.

\begin{table*}[t]
\caption{Zero-shot benchmark results on MagicGarment. U1/U2 denote ULIP-1/2, Uni denotes Uni3D, and I/T denote image/text similarity.}
\label{tab:benchmark-all}
\centering
\setlength{\tabcolsep}{0.5pt}
\begin{tabular*}{\linewidth}{@{}l@{\extracolsep{\fill}}*{13}{c}@{}}
\toprule
& \multicolumn{2}{c}{Geometry}
& \multicolumn{3}{c}{Appearance}
& \multicolumn{2}{c}{Normal}
& \multicolumn{6}{c}{Semantic} \\
\cmidrule(lr){2-3}\cmidrule(lr){4-6}\cmidrule(lr){7-8}\cmidrule(l){9-14}
Method & CD$\downarrow$ & F-score$\uparrow$ & PSNR$\uparrow$ & SSIM$\uparrow$ & LPIPS$\downarrow$ & PSNR-N$\uparrow$ & LPIPS-N$\downarrow$ & U1-I$\uparrow$ & U1-T$\uparrow$ & U2-I$\uparrow$ & U2-T$\uparrow$ & Uni-I$\uparrow$ & Uni-T$\uparrow$ \\
\midrule
Hunyuan3D-2.1 & \textbf{4.2322} & 12.5148 & 17.6060 & 0.8193 & \textbf{0.2294} & 16.0559 & 0.2386 & 0.1139 & 0.0612 & \textbf{0.3975} & \textbf{0.2845} & \textbf{0.3411} & \textbf{0.2601} \\
AIpparel       & 34.4063 & 5.0719  & 14.1433 & 0.7452 & 0.4192 & 9.5482  & 0.4608 & 0.0613 & -0.0336 & 0.1276 & 0.0731 & 0.0837 & 0.0952 \\
ChatGarment    & 39.4414 & 4.6322  & 14.1324 & 0.7617 & 0.4161 & 9.8198  & 0.4453 & 0.0607 & -0.0324 & 0.1203 & 0.0696 & 0.0899 & 0.1046 \\
CRM            & 11.5174 & 6.5935  & \textbf{18.5177} & 0.8252 & 0.2431 & 15.3413 & 0.2646 & 0.1253 & \textbf{0.0688} & 0.3078 & 0.2043 & 0.2712 & 0.1966 \\
LGM            & 16.5124 & 5.2953  & 18.1696 & \textbf{0.8254} & 0.2531 & 15.5809 & 0.2594 & \textbf{0.1392} & 0.0599 & 0.2844 & 0.1855 & 0.2487 & 0.1987 \\
One-2-3-45     & 31.4587 & 3.8417  & 12.7375 & 0.7986 & 0.3299 & 14.3821 & 0.2976 & 0.1177 & 0.0229 & 0.2248 & 0.1441 & 0.1868 & 0.1753 \\
InstantMesh    & 14.0708 & 6.2238  & 15.8091 & 0.8218 & 0.2402 & \textbf{16.1878} & \textbf{0.2279} & 0.1376 & 0.0601 & 0.3778 & 0.2604 & 0.3141 & 0.2479 \\
TRELLIS        & 5.9640  & 10.1691 & 16.4500 & 0.7172 & 0.3411 & 12.0796 & 0.3667 & 0.1240 & 0.0579 & 0.3557 & 0.2440 & 0.3030 & 0.2213 \\
TRELLIS.2      & 4.5291  & \textbf{13.6206} & 16.1205 & 0.7597 & 0.3024 & 13.0309 & 0.3207 & 0.1313 & 0.0612 & 0.3526 & 0.2368 & 0.2973 & 0.2172 \\
Unique3D       & 17.1257 & 6.5459  & 12.3284 & 0.5479 & 0.5425 & 8.4922  & 0.5804 & 0.1266 & 0.0583 & 0.3135 & 0.2077 & 0.2532 & 0.1796 \\
Wonder3D       & 13.7902 & 7.7564  & 13.3405 & 0.8126 & 0.2984 & 15.3782 & 0.2583 & 0.1381 & 0.0324 & 0.3566 & 0.2493 & 0.3030 & 0.2408 \\
\bottomrule
\end{tabular*}
\vspace{-1.5em}
\end{table*}

\begin{table*}[t]
\caption{In-domain adaptation results on MagicGarment. U1/U2 denote ULIP-1/2, Uni denotes Uni3D, and I/T denote image/text similarity.}
\label{tab:trained-benchmark}
\centering
\setlength{\tabcolsep}{0.5pt}
\begin{tabular*}{\linewidth}{@{}l@{\extracolsep{\fill}}*{13}{c}@{}}
\toprule
& \multicolumn{2}{c}{Geometry}
& \multicolumn{3}{c}{Appearance}
& \multicolumn{2}{c}{Normal}
& \multicolumn{6}{c}{Semantic} \\
\cmidrule(lr){2-3}\cmidrule(lr){4-6}\cmidrule(lr){7-8}\cmidrule(l){9-14}
Method & CD$\downarrow$ & F-score$\uparrow$ & PSNR$\uparrow$ & SSIM$\uparrow$ & LPIPS$\downarrow$ & PSNR-N$\uparrow$ & LPIPS-N$\downarrow$ & U1-I$\uparrow$ & U1-T$\uparrow$ & U2-I$\uparrow$ & U2-T$\uparrow$ & Uni-I$\uparrow$ & Uni-T$\uparrow$ \\
\midrule
Hunyuan3D-2.1 & 2.7026 & 14.0527 & 19.0146 & 0.8380 & 0.1889 & 17.6048 & 0.2069 & 0.1162 & 0.0597 & 0.3900 & 0.2813 & \textbf{0.3475} & 0.2584 \\
InstantMesh   & 11.3486 & 8.0174 & 17.6216 & 0.8475 & 0.1962 & 18.0185 & 0.2140 & \textbf{0.1402} & 0.0586 & 0.3751 & 0.2657 & 0.3198 & 0.2455 \\
TRELLIS       & 4.3491 & 11.6548 & 17.9737 & 0.7390 & 0.2810 & 13.6614 & 0.2659 & 0.1227 & 0.0592 & 0.3629 & 0.2418 & 0.3094 & 0.2199 \\
TRELLIS.2     & 3.2618 & \textbf{15.2347} & 17.8024 & 0.7824 & 0.2518 & 14.6542 & 0.2292 & 0.1334 & \textbf{0.0599} & 0.3588 & 0.2342 & 0.3035 & 0.2156 \\
Wonder3D      & 11.0617 & 10.4935 & 16.0713 & 0.8296 & 0.2480 & 16.6064 & 0.2332 & 0.1359 & 0.0340 & 0.3642 & 0.2461 & 0.3008 & 0.2467 \\
\textbf{OURS}  & \textbf{1.8325} & 9.0500 & \textbf{21.1095} & \textbf{0.8593} & \textbf{0.1216} & \textbf{21.7626} & \textbf{0.0949} & 0.1191 & 0.0555 & \textbf{0.4041} & \textbf{0.2874} & 0.3454 & \textbf{0.2651} \\
\bottomrule
\end{tabular*}
\vspace{-1.5em}
\end{table*}

\textbf{Flow-Time-Route LoRA (FTR-LoRA).}
Built upon the pretrained Hunyuan3D-DiT backbone, FTR-LoRA performs lightweight garment-domain adaptation while preserving its generic 3D generative priors. Most backbone parameters are frozen, and adaptation is restricted to low-rank branches and lightweight controller-related parameters. Unlike standard LoRA~\cite{Hu2022LoRA}, which applies a static low-rank residual, FTR-LoRA dynamically modulates individual rank channels using the flow timestep, token-level routing state, and sample-level expert-load statistics. We adopt the rank-stabilized scaling of rsLoRA~\cite{Kalajdzievski2023RankStabilization}.

For token representation $h_n$, the MoE router predicts expert probabilities $\pi_n$ and the Top-$K$ expert set $\mathcal{E}_n$. We summarize token-level routing and sample-level expert usage as
\begin{equation}
\begin{aligned}
\pi_n &= \operatorname{softmax}(W_g h_n), \qquad
\mathcal{E}_n=\operatorname{TopK}(\pi_n),\\
R_n &= [\pi_n,\mathcal{H}(\pi_n),
        \pi_{n,(1)}-\pi_{n,(2)}],\\
L_b &= [\bar{\pi}_b,\bar{e}_b], \qquad
\bar{\pi}_b=\frac{1}{N}\sum_{m=1}^{N}\pi_m,\quad
\bar{e}_b=\frac{1}{N}\sum_{m=1}^{N}e_m .
\end{aligned}
\label{eq:ftr_context}
\end{equation}
where $\mathcal{H}(\cdot)$ denotes routing entropy, $\pi_{n,(1)}$ and $\pi_{n,(2)}$ are the two largest routing probabilities, and $e_m$ is the multi-hot indicator of the experts selected for token $m$.

Given the flow timestep $t$, token-level route feature $R_n$, and sample-level expert-load feature $L_b$, a lightweight controller generates the rank-wise gate:
\begin{equation}
\begin{aligned}
c_{\ell,n} &=
[\phi_t(t),\phi_r(R_n),\phi_e(L_b)],\\
\Gamma_{\ell,n}
&=
\sqrt{r_\ell}\,
\frac{
1+\beta\tanh\!\left(U_\ell\Phi_\ell(c_{\ell,n})\right)}
{\left\|
1+\beta\tanh\!\left(U_\ell\Phi_\ell(c_{\ell,n})\right)
\right\|_2+\epsilon},
\end{aligned}
\label{eq:ftr_gate}
\end{equation}
where $\Phi_\ell$ and $U_\ell$ denote the shared controller and layer-specific head, respectively, and $r_\ell$ is the LoRA rank. For a frozen target layer $W_\ell$, FTR-LoRA computes
\begin{equation}
y_{\ell,n}
=
W_\ell h_n+
\frac{\alpha_\ell}{\sqrt{r_\ell}}
B_\ell
\left(
\Gamma_{\ell,n}\odot A_\ell h_n
\right),
\label{eq:ftr_forward}
\end{equation}
where $A_\ell$ and $B_\ell$ are the trainable low-rank projections and $\alpha_\ell$ is the scaling coefficient. The dynamic gate therefore adjusts rank-wise adaptation according to both the flow stage and routing context, enabling the pretrained 3D prior to better accommodate thin shells, open boundaries, multilayer structures, and fine garment details.

\section{Experiments}\label{sec:experiments}

\subsection{Experimental Setup}
\label{subsec:experimental-setup}

\textbf{Baselines.}
We benchmark representative garment-specific and general-purpose image-to-3D reconstruction methods, including AIpparel~\cite{Nakayama2025AIpparel}, ChatGarment~\cite{Bian2025ChatGarment}, InstantMesh~\cite{Xu2024InstantMesh}, CRM~\cite{Wang2024CRM}, LGM~\cite{Tang2024LGM}, One-2-3-45~\cite{liu2023one2345}, Unique3D~\cite{Wu2024Unique3D}, Wonder3D~\cite{Long2024Wonder3D}, Hunyuan3D-\allowbreak{}2.1 \cite{hunyuan3d2025hunyuan3d}, TRELLIS~\cite{Xiang2025TRELLIS}, and TRELLIS.2~\cite{Xiang2025TRELLIS2}. 
AIpparel and ChatGarment operate within predefined garment spaces without general texture generation, but are included as garment-specific baselines.
All outputs are evaluated under the unified protocol in Sec.~\ref{subsec:benchmark}.

\textbf{Metrics.}
Geometry is evaluated using Chamfer Distance (CD) and F-score with a threshold of $0.01$. Appearance fidelity is measured by multi-view PSNR, SSIM, and LPIPS, while surface orientation is evaluated using PSNR-N and LPIPS-N on rendered normal maps. We additionally report image--point-cloud and text--point-cloud similarities using ULIP, ULIP-2, and Uni3D~\cite{Xue2023ULIP,Xue2024ULIP2,Zhou2024Uni3D} to evaluate cross-modal semantic consistency.
All metrics are averaged over the test set.

\textbf{Implementation.}
Our reference baseline is trained using AdamW with a learning rate of $1\times10^{-5}$ for 700K steps on four NVIDIA A30 GPUs. At inference, we use 50-step Euler ODE sampling for shape-latent generation.





\subsection{Zero-Shot Benchmark}
\label{subsec:zero-shot}

Table~\ref{tab:benchmark-all} reports the zero-shot results on MagicGarment. Existing methods exhibit clear trade-offs across different evaluation dimensions. Hunyuan3D-2.1 achieves the lowest CD (4.2322) and LPIPS (0.2294), together with the strongest performance on most ULIP-2 and Uni3D metrics, while TRELLIS.2 obtains the highest F-score (13.6206). CRM performs best in RGB appearance fidelity with a PSNR of 18.5177 and an SSIM of 0.8252, whereas InstantMesh achieves the strongest normal-map reconstruction with a PSNR-N of 16.1878 and an LPIPS-N of 0.2279. These results show that no single method consistently dominates all evaluation dimensions, highlighting the complementary challenges captured by MagicGarment.


\subsection{In-Domain Adaptation}
\label{subsec:adaptation}

Table~\ref{tab:trained-benchmark} reports the results after adaptation on the MagicGarment training set. Our garment-aware baseline achieves the best performance on 9 of 13 metrics, including CD, PSNR, SSIM, LPIPS, both surface-orientation metrics, ULIP-2 image/text similarity, and Uni3D text similarity. Compared with the adapted Hunyuan3D-2.1 backbone, our method reduces CD from 2.7026 to 1.8325 and LPIPS from 0.1889 to 0.1216, while improving PSNR from 19.0146 to 21.1095 and PSNR-N from 17.6048 to 21.7626. These gains indicate that the proposed garment-specific adaptation substantially improves geometric fidelity, appearance consistency, and surface reconstruction. Meanwhile, TRELLIS.2 achieves the highest F-score, suggesting that different reconstruction objectives still exhibit complementary strengths.

\textbf{Qualitative Comparison.}
As shown in Fig.~\ref{fig:qualitative-results}, existing methods exhibit visible structural errors around garment openings, thin surfaces, and layered regions. Our garment-aware baseline more consistently preserves the garment silhouette, open boundaries, and local structural details across diverse examples, supporting the quantitative improvements obtained through garment-specific adaptation.



\section{Conclusion}


This work introduces MagicGarment, a benchmark for single-image standalone 3D garment reconstruction. Our evaluation reveals substantial performance differences among existing methods, highlighting the distinct challenges posed by garment-specific geometry and the limitations of directly transferring general-purpose 3D reconstruction models. The proposed garment-aware reference baseline further shows the value of domain-specific adaptation. Future work will extend MagicGarment with greater scale, diversity, and more challenging real-world scenarios.

\section{Compliance with Ethical Standards}\label{sec:ethics}

This study involves no human participants or animal subjects. The human body models are synthetic template-based assets used solely for garment simulation; therefore, no ethical approval was required.

\section{Acknowledgements}\label{sec:acknowledgements}

This work was supported by the National Natural Science Foundation of China under Grant No. 62202345.

\bibliographystyle{IEEEbib}
\bibliography{strings,refs}

@inproceedings{Long2024Wonder3D,
  author    = {Long, Xiaoxiao and Guo, Yuan-Chen and Lin, Cheng and Liu, Yuan and Dou, Zhiyang and Liu, Lingjie and Ma, Yuexin and Zhang, Song-Hai and Habermann, Marc and Theobalt, Christian and Wang, Wenping},
  title     = {{Wonder3D}: Single Image to {3D} Using Cross-Domain Diffusion},
  fullbooktitle = {Proceedings of the IEEE/CVF Conference on Computer Vision and Pattern Recognition},
  booktitle = {CVPR},
  pages     = {9970--9980},
  year      = {2024},
  doi       = {10.1109/CVPR52733.2024.00951}
}

@inproceedings{Tang2024LGM,
  author    = {Tang, Jiaxiang and Chen, Zhaoxi and Chen, Xiaokang and Wang, Tengfei and Zeng, Gang and Liu, Ziwei},
  title     = {{LGM}: Large Multi-View Gaussian Model for High-Resolution {3D} Content Creation},
  fullbooktitle = {European Conference on Computer Vision},
  booktitle = {ECCV},
  pages     = {1--18},
  year      = {2024},
  doi       = {10.1007/978-3-031-73235-5_1}
}

@inproceedings{Xiang2025TRELLIS,
  author    = {Xiang, Jianfeng and Lv, Zelong and Xu, Sicheng and Deng, Yu and Wang, Ruicheng and Zhang, Bowen and Chen, Dong and Tong, Xin and Yang, Jiaolong},
  title     = {Structured {3D} Latents for Scalable and Versatile {3D} Generation},
  fullbooktitle = {Proceedings of the IEEE/CVF Conference on Computer Vision and Pattern Recognition},
  booktitle = {CVPR},
  pages     = {21469--21480},
  year      = {2025},
  doi       = {10.1109/CVPR52734.2025.02000}
}

@inproceedings{Jiang2020BCNet,
  author    = {Jiang, Boyi and Zhang, Juyong and Hong, Yang and Luo, Jinhao and Liu, Ligang and Bao, Hujun},
  title     = {{BCNet}: Learning Body and Cloth Shape from a Single Image},
  fullbooktitle = {European Conference on Computer Vision},
  booktitle = {ECCV},
  pages     = {18--35},
  year      = {2020},
  doi       = {10.1007/978-3-030-58565-5_2}
}

@inproceedings{DeLuigi2023DrapeNet,
  author    = {De Luigi, Luca and Li, Ren and Guillard, Beno{\^i}t and Salzmann, Mathieu and Fua, Pascal},
  title     = {{DrapeNet}: Garment Generation and Self-Supervised Draping},
  fullbooktitle = {Proceedings of the IEEE/CVF Conference on Computer Vision and Pattern Recognition},
  booktitle = {CVPR},
  pages     = {1451--1460},
  year      = {2023},
  doi       = {10.1109/CVPR52729.2023.00146}
}

@inproceedings{Bhatnagar2019MGN,
  author    = {Bhatnagar, Bharat Lal and Tiwari, Garvita and Theobalt, Christian and Pons-Moll, Gerard},
  title     = {Multi-Garment Net: Learning to Dress {3D} People From Images},
  fullbooktitle = {Proceedings of the IEEE/CVF International Conference on Computer Vision},
  booktitle = {ICCV},
  pages     = {5420--5430},
  year      = {2019},
  doi       = {10.1109/ICCV.2019.00552}
}

@inproceedings{Zhu2020DeepFashion3D,
  author    = {Zhu, Heming and Cao, Yu and Jin, Hang and Chen, Weikai and Du, Dong and Wang, Zhangye and Cui, Shuguang and Han, Xiaoguang},
  title     = {{Deep Fashion3D}: A Dataset and Benchmark for {3D} Garment Reconstruction from Single Images},
  fullbooktitle = {European Conference on Computer Vision},
  booktitle = {ECCV},
  pages     = {512--530},
  year      = {2020},
  doi       = {10.1007/978-3-030-58452-8_30}
}

@inproceedings{Bertiche2020CLOTH3D,
  author    = {Bertiche, Hugo and Madadi, Meysam and Escalera, Sergio},
  title     = {{CLOTH3D}: Clothed {3D} Humans},
  fullbooktitle = {European Conference on Computer Vision},
  booktitle = {ECCV},
  pages     = {344--359},
  year      = {2020},
  doi       = {10.1007/978-3-030-58565-5_21}
}

@inproceedings{Wang2024FourDDress,
  author    = {Wang, Wenbo and Ho, Hsuan-I and Guo, Chen and Rong, Boxiang and Grigorev, Artur and Song, Jie and Zarate, Juan Jose and Hilliges, Otmar},
  title     = {{4D-DRESS}: A {4D} Dataset of Real-World Human Clothing with Semantic Annotations},
  fullbooktitle = {Proceedings of the IEEE/CVF Conference on Computer Vision and Pattern Recognition},
  booktitle = {CVPR},
  pages     = {550--560},
  year      = {2024},
  doi       = {10.1109/CVPR52733.2024.00059}
}

@article{Luo2024GarVerseLOD,
  author    = {Luo, Zhongjin and Liu, Haolin and Li, Chenghong and Du, Wanghao and Jin, Zirong and Sun, Wanhu and Nie, Yinyu and Chen, Weikai and Han, Xiaoguang},
  title     = {{GarVerseLOD}: High-Fidelity {3D} Garment Reconstruction from a Single In-the-Wild Image using a Dataset with Levels of Details},
  journal   = {ACM Transactions on Graphics},
  volume    = {43},
  number    = {6},
  pages     = {1--12},
  year      = {2024},
  publisher = {Association for Computing Machinery},
  doi       = {10.1145/3687921}
}

@misc{CLO3D2026,
  author       = {{CLO Virtual Fashion}},
  title        = {{CLO}: {3D} Fashion Design Software},
  year         = {2026},
  howpublished = {\url{https://www.clo3d.com/en/}},
  note         = {Accessed: 2026-09-24}
}

@inproceedings{liu2024grounding,
  author    = {Liu, Shilong and Zeng, Zhaoyang and Ren, Tianhe and Li, Feng and Zhang, Hao and Yang, Jie and Jiang, Qing and Li, Chunyuan and Yang, Jianwei and Su, Hang and Zhu, Jun and Zhang, Lei},
  title     = {{Grounding DINO}: Marrying {DINO} with Grounded Pre-Training for Open-Set Object Detection},
  fullbooktitle = {European Conference on Computer Vision},
  booktitle = {ECCV},
  pages     = {38--55},
  year      = {2024},
  doi       = {10.1007/978-3-031-72970-6_3}
}

@inproceedings{Li2024SemanticSAM,
  author    = {Li, Feng and Zhang, Hao and Sun, Peize and Zou, Xueyan and Liu, Shilong and Li, Chunyuan and Yang, Jianwei and Zhang, Lei and Gao, Jianfeng},
  title     = {Segment and Recognize Anything at Any Granularity},
  fullbooktitle = {European Conference on Computer Vision},
  booktitle = {ECCV},
  pages     = {467--484},
  year      = {2024},
  doi       = {10.1007/978-3-031-73195-2_27}
}

@inproceedings{kim2025zim,
  author    = {Kim, Beomyoung and Shin, Chanyong and Jeong, Joonhyun and Jung, Hyungsik and Lee, Se-Yun and Chun, Sewhan and Hwang, Dong-Hyun and Yu, Joonsang},
  title     = {{ZIM}: Zero-Shot Image Matting for Anything},
  fullbooktitle = {Proceedings of the IEEE/CVF International Conference on Computer Vision},
  booktitle = {ICCV},
  pages     = {23828--23838},
  year      = {2025},
  doi       = {10.1109/ICCV51701.2025.02212}
}

@inproceedings{Hu2022LoRA,
  author    = {Hu, Edward J. and Shen, Yelong and Wallis, Phillip and Allen-Zhu, Zeyuan and Li, Yuanzhi and Wang, Shean and Wang, Lu and Chen, Weizhu},
  title     = {{LoRA}: Low-Rank Adaptation of Large Language Models},
  fullbooktitle = {International Conference on Learning Representations},
  booktitle = {ICLR},
  year      = {2022}
}

@article{Kalajdzievski2023RankStabilization,
  author  = {Kalajdzievski, Damjan},
  title   = {A Rank Stabilization Scaling Factor for Fine-Tuning with {LoRA}},
  journal = {arXiv preprint arXiv:2312.03732},
  year    = {2023},
  doi     = {10.48550/arXiv.2312.03732}
}

@inproceedings{Nakayama2025AIpparel,
  author    = {Nakayama, Kiyohiro and Ackermann, Jan and Kesdogan, Timur Levent and Zheng, Yang and Korosteleva, Maria and Sorkine-Hornung, Olga and Guibas, Leonidas J. and Yang, Guandao and Wetzstein, Gordon},
  title     = {{AIpparel}: A Multimodal Foundation Model for Digital Garments},
  fullbooktitle = {Proceedings of the IEEE/CVF Conference on Computer Vision and Pattern Recognition},
  booktitle = {CVPR},
  pages     = {8138--8149},
  year      = {2025},
  doi       = {10.1109/CVPR52734.2025.00762}
}

@inproceedings{Bian2025ChatGarment,
  author    = {Bian, Siyuan and Xu, Chenghao and Xiu, Yuliang and Grigorev, Artur and Liu, Zhen and Lu, Cewu and Black, Michael J. and Feng, Yao},
  title     = {{ChatGarment}: Garment Estimation, Generation and Editing via Large Language Models},
  fullbooktitle = {Proceedings of the IEEE/CVF Conference on Computer Vision and Pattern Recognition},
  booktitle = {CVPR},
  pages     = {2924--2934},
  year      = {2025},
  doi       = {10.1109/CVPR52734.2025.00278}
}

@article{Xu2024InstantMesh,
  author  = {Xu, Jiale and Cheng, Weihao and Gao, Yiming and Wang, Xintao and Gao, Shenghua and Shan, Ying},
  title   = {{InstantMesh}: Efficient {3D} Mesh Generation from a Single Image with Sparse-View Large Reconstruction Models},
  journal = {arXiv preprint arXiv:2404.07191},
  year    = {2024}
}

@inproceedings{Wang2024CRM,
  author    = {Wang, Zhengyi and Wang, Yikai and Chen, Yifei and Xiang, Chendong and Chen, Shuo and Yu, Dajiang and Li, Chongxuan and Su, Hang and Zhu, Jun},
  title     = {{CRM}: Single Image to {3D} Textured Mesh with Convolutional Reconstruction Model},
  fullbooktitle = {European Conference on Computer Vision},
  booktitle = {ECCV},
  pages     = {57--74},
  year      = {2024},
  doi       = {10.1007/978-3-031-72751-1_4}
}

@article{liu2023one2345,
  title   = {One-2-3-45: Any single image to 3d mesh in 45 seconds without per-shape optimization},
  author  = {Liu, Minghua and Xu, Chao and Jin, Haian and Chen, Linghao and Varma T, Mukund and Xu, Zexiang and Su, Hao},
  journal = {Advances in Neural Information Processing Systems},
  volume  = {36},
  year    = {2024}
}

@inproceedings{Wu2024Unique3D,
  author    = {Wu, Kailu and Liu, Fangfu and Cai, Zhihan and Yan, Runjie and Wang, Hanyang and Hu, Yating and Duan, Yueqi and Ma, Kaisheng},
  title     = {{Unique3D}: High-Quality and Efficient {3D} Mesh Generation from a Single Image},
  fullbooktitle = {Advances in Neural Information Processing Systems},
  booktitle = {NeurIPS},
  volume    = {37},
  pages     = {125116--125141},
  year      = {2024},
  doi       = {10.52202/079017-3974}
}

@misc{hunyuan3d2025hunyuan3d,
  author        = {{Tencent Hunyuan3D Team}},
  title         = {{Hunyuan3D 2.1}: From Images to High-Fidelity {3D} Assets with Production-Ready {PBR} Material},
  year          = {2025},
  howpublished  = {arXiv preprint arXiv:2506.15442},
  eprint        = {2506.15442},
  archivePrefix = {arXiv},
  primaryClass  = {cs.CV}
}

@inproceedings{Xiang2025TRELLIS2,
  author    = {Xiang, Jianfeng and Chen, Xiaoxue and Xu, Sicheng and Wang, Ruicheng and Lv, Zelong and Deng, Yu and Zhu, Hongyuan and Dong, Yue and Zhao, Hao and Yuan, Nicholas Jing and Yang, Jiaolong},
  title     = {Native and Compact Structured Latents for {3D} Generation},
  fullbooktitle = {Proceedings of the IEEE/CVF Conference on Computer Vision and Pattern Recognition},
  booktitle = {CVPR},
  pages     = {14419--14429},
  year      = {2026}
}

@inproceedings{Xue2023ULIP,
  author    = {Xue, Le and Gao, Mingfei and Xing, Chen and Mart{\'i}n-Mart{\'i}n, Roberto and Wu, Jiajun and Xiong, Caiming and Xu, Ran and Niebles, Juan Carlos and Savarese, Silvio},
  title     = {{ULIP}: Learning a Unified Representation of Language, Images, and Point Clouds for {3D} Understanding},
  fullbooktitle = {Proceedings of the IEEE/CVF Conference on Computer Vision and Pattern Recognition},
  booktitle = {CVPR},
  pages     = {1179--1189},
  year      = {2023},
  doi       = {10.1109/CVPR52729.2023.00120}
}

@inproceedings{Xue2024ULIP2,
  author    = {Xue, Le and Yu, Ning and Zhang, Shu and Panagopoulou, Artemis and Li, Junnan and Mart{\'i}n-Mart{\'i}n, Roberto and Wu, Jiajun and Xiong, Caiming and Xu, Ran and Niebles, Juan Carlos and Savarese, Silvio},
  title     = {{ULIP-2}: Towards Scalable Multimodal Pre-Training for {3D} Understanding},
  fullbooktitle = {Proceedings of the IEEE/CVF Conference on Computer Vision and Pattern Recognition},
  booktitle = {CVPR},
  pages     = {27091--27101},
  year      = {2024},
  doi       = {10.1109/CVPR52733.2024.02562}
}

@inproceedings{Zhou2024Uni3D,
  author    = {Zhou, Junsheng and Wang, Jinsheng and Ma, Baorui and Liu, Yu-Shen and Huang, Tiejun and Wang, Xinlong},
  title     = {{Uni3D}: Exploring Unified {3D} Representation at Scale},
  fullbooktitle = {International Conference on Learning Representations},
  booktitle = {ICLR},
  year      = {2024},
  url       = {https://openreview.net/forum?id=wcaE4Dfgt8}
}

\end{document}